\documentclass[a4paper,UKenglish]{lipics-v2016}
\newif\ifDoubleBlind
\DoubleBlindtrue

\usepackage{amsfonts}

\newcommand{\set}[1]{\left\{ #1\right\}}

\newcommand{\sodass}{\,:\,}
\newcommand{\setGilt}[2]{\left\{ #1\sodass #2\right\}}

\newcommand{\realrange}[2]{\left[#1, #2\right]}

\newcommand{\unitrange}[2]{\realrange{0}{1}}

\newcommand{\Oh}[1]{\mathcal{O}\!\left( #1\right)}

\newcommand{\llabel}[1]{\label{\labelprefix:#1}}
\newcommand{\labelprefix}{} 

\newcommand{\discussionsize}{\small}

\newcommand{\frage}[1]{}

\newenvironment{code}{\noindent
\begin{tabbing}%
\hspace{2em}\=\hspace{2em}\=\hspace{2em}\=\hspace{2em}\=\hspace{2em}\=%
\hspace{2em}\=\hspace{2em}\=\hspace{2em}\=\hspace{2em}\=\hspace{2em}\=%
\kill}{\end{tabbing}}

\newcommand{\labelcommand}{}
\newcommand{\captiontext}{}
\newsavebox{\codeparam}
\newcounter{lineNumber}
\newenvironment{disscodepos}[3]{%
\renewcommand{\labelcommand}{#2}%
\renewcommand{\captiontext}{#3}%
\sbox{\codeparam}{\parbox{\textwidth}{#3}}%
\begin{figure}[#1]\begin{center}\begin{code}\setcounter{lineNumber}{1}}{%
\end{code}\end{center}\caption{\llabel{\labelcommand}\captiontext}\end{figure}}

{\end{disscodepos}}

\newcommand{\Is}       {:=}

\newdimen\endofsize\endofsize=0.5em
\def\endofbeweis{~\quad\hglue\hsize minus\hsize
                 \hbox{\vrule height \endofsize width
\endofsize}\par}

\usepackage{multirow}
\usepackage{algorithmic}
\usepackage{algorithm}
\usepackage{xspace}
\usepackage{longtable}
\usepackage{numprint}
\usepackage{wrapfig}

\usepackage{amsmath}
\usepackage{hyperref}
\usepackage[usenames,dvipsnames]{xcolor}
\usepackage[normalem]{ulem}
\usepackage[]{graphicx}\usepackage[]{color}
\makeatletter
\def\maxwidth{ %
  \ifdim\Gin@nat@width>\linewidth
    \linewidth
  \else
    \Gin@nat@width
  \fi
}
\makeatother

\newcommand{\term}[1]{\textsl{#1}}

\newcommand{\cov}{\term{cov}\xspace}
\definecolor{fgcolor}{rgb}{0.345, 0.345, 0.345}

\usepackage{framed}
\makeatletter
 {\par\unskip\endMakeFramed%
 \at@end@of@kframe}
\makeatother

\definecolor{shadecolor}{rgb}{.97, .97, .97}
\definecolor{messagecolor}{rgb}{0, 0, 0}
\definecolor{warningcolor}{rgb}{1, 0, 1}
\definecolor{errorcolor}{rgb}{1, 0, 0}

\usepackage{alltt}
\newcommand{\SweaveOpts}[1]{}  
\newcommand{\SweaveInput}[1]{} 
\newcommand{\Sexpr}[1]{}       

\newcommand{\ie}{i.e.~}
\newcommand{\etal}{et~al.\ }

\usepackage{capt-of}%
\usepackage{caption}

\usepackage{etoolbox}
\usepackage[left,pagewise] {lineno}
\definecolor {infocolor} {rgb} {0.6,0.6,0.6}

\patchcmd{\thebibliography}{\list}{\fontsize{0.98em}{0.9\baselineskip}\selectfont\list}{}{}

\newcommand{\mytitle}{Memetic Graph Clustering}
\begin{document}
\title{\mytitle}
\author[1]{Sonja Biedermann}
\author[2]{Monika Henzinger}
\author[3]{Christian Schulz}
\author[4]{Bernhard Schuster}
\affil[1]{University of Vienna, Vienna, Austria\\
  \texttt{sonja.biedermann@univie.ac.at}}
\affil[2]{University of Vienna, Vienna, Austria\\
  \texttt{monika.henzinger@univie.ac.at}}
\affil[3]{University of Vienna, Vienna, Austria\\
  \texttt{christian.schulz@univie.ac.at}}
\affil[4]{University of Vienna, Vienna, Austria\\
  \texttt{bernhard.schuster@univie.ac.at}}
\authorrunning{S. Biedermann et. al.} 


\subjclass{G.2.2 Graph Theory}
\keywords{Graph Clustering, Evolutionary Algorithms}

\EventLongTitle{}
\EventShortTitle{}
\EventAcronym{}
\EventYear{2018}
\EventDate{}

\EventLocation{}
\EventLogo{}

\maketitle
\begin{abstract}
It is common knowledge that there is no single best strategy for graph clustering, which justifies a plethora of existing approaches.
In this paper, we present a general memetic algorithm, VieClus, to tackle the graph clustering problem. This algorithm can be adapted to optimize different objective functions.
A key component of our contribution are natural recombine operators that employ ensemble clusterings as well as multi-level techniques. 
Lastly, we combine these techniques with a scalable communication protocol, producing a system that is able to compute high-quality solutions in a short amount of time.
We instantiate our scheme with local search for modularity and show that our algorithm successfully improves or reproduces all entries of the 10th DIMACS implementation~challenge under consideration using a small amount of time.

\end{abstract}

\section{Introduction}
Graph clustering is the problem of detecting tightly connected regions of a
graph. Depending on the task, knowledge about the structure of the graph can
reveal information such as voter behavior, the formation of new trends, existing
terrorist groups and recruitment~\cite{survey} or a natural partitioning of
data records onto pages~\cite{cluster-paging}. Further application areas
include the study of protein interaction~\cite{cluster-protein}, gene
expression networks~\cite{cluster-geneexp}, fraud
detection~\cite{cluster-anomalies}, program optimization~\cite{cluster-opt1,cluster-opt2} and the spread of
epidemics~\cite{cluster-epidemic}---possible applications are plentiful, as
almost all systems containing interacting or coexisting entities can be modeled
as a graph.

It is common knowledge that there is no single best strategy for graph clustering, which justifies a plethora of existing approaches.
Moreover, most quality indices for graph clusterings have turned out to be NP-hard to optimize and are rather resilient to effective approximations, see e.g. \cite{ausiello2012complexity,brandes2008modularity,wagner1993between}, allowing only heuristic approaches for optimization.
The majority of algorithms for graph clustering are based on the paradigm of intra-cluster density versus inter-cluster sparsity.
One successful heuristic to cluster large graphs is the \emph{multi-level} approach~\cite{brandes2005network}, e.g. the Louvain method for the optimization of modularity~\cite{louvain}.
Here, the graph is recursively \emph{contracted} to obtain smaller graphs which should reflect the same general structure as the input. 
After applying an \emph{initial clustering} algorithm to the smallest graph, the contraction steps are undone and, at each level, a
\emph{local search} method is used to improve the clustering induced by the coarser level w.r.t some objective function measuring the quality of the clustering. 
The intuition behind this approach is that a good clustering at one level of the hierarchy will also be a good clustering on the next finer level. Hence, 
depending on the definition of the neighborhood, local search algorithms are able to explore local solution spaces very effectively~in~this~setting. 
However, these methods are also prone to get trapped in local optima. 
The multi-level scheme can help to some extent since 
local search has a more global view on the problem on the coarse levels and a very fine-grained view on the fine levels of the multi-level hierarchy. 
In addition, as with many other randomized meta-heuristics, several repeated runs can be made in order to improve the final result at the expense of running time.

Still, even a large number of repeated executions can only scratch the surface of the huge search space of possible clusterings. 
In order to explore the global solution space extensively, we need more sophisticated meta-heuristics. 
This is where memetic algorithms (MAs), \ie genetic algorithms combined with local search~\cite{KimHKM11}, 
come into play. Memetic algorithms allow for effective exploration (global search) and exploitation (local search) of the solution space.
The general idea behind genetic algorithms is to use mechanisms inspired by biological evolution such as selection, mutation, recombination, and survival of the fittest. 
A genetic algorithm (GA) starts with a population of individuals (in our case clusterings of the graph) and evolves the population over several generational cycles~(rounds).
In each round, the GA uses a selection rule based on the fitness of the individuals of the population to select good individuals and combines them to obtain improved offspring~\cite{goldbergGA89}. 
When an offspring is generated an eviction rule is used to select a member of the population to be replaced by the new offspring. 
For an evolutionary algorithm it is of major importance to preserve diversity in the population~\cite{baeckEvoAlgPHD96}, i.e., the individuals should not become too similar in order to avoid a premature convergence of the~algorithm.
This is usually achieved by using mutation operations and by using eviction rules that take similarity of individuals~into~account.

In this paper, we present a general memetic algorithm, VieClus (Vienna Graph Clustering), to tackle the graph clustering problem. This algorithm can be adapted to optimize different objective functions simply by using a local search algorithm that optimizes the objective function desired by the user.
A key component of our contribution are natural recombine operators that employ ensemble clusterings as well as multi-level techniques. 
In machine learning, ensemble methods combine multiple weak classification (or clustering) algorithms to obtain a strong algorithm for classification (or clustering).
More precisely, given a number of clusterings, the \emph{overlay/ensemble clustering} is a clustering in which two vertices belong to the same cluster if and only if they belong to the same cluster in each of the input clusterings. 
Our recombination operators use the overlay of two clusterings from the population to decide whether pairs of vertices should belong to the same cluster~\cite{OvelgoenneG13ensemble,staudtmeyerhenke13high}.
This is combined with a local search algorithm to find further improvements and also embedded into a multi-level algorithm to find even better clusterings.
Our general principle is to randomize tie-breaking whenever possible. This diversifies the search and also improves solutions.
Lastly, we combine these techniques with a scalable communication protocol, producing a system that is able to compute high-quality solutions in a short amount of time.
In our experimental evaluation, we show that our algorithm successfully improves or reproduces all entries of the 10th DIMACS implementation~challenge under consideration in a small amount of time. In fact, for most of the small instances, we can improve the old benchmark result \emph{in less than a minute}.
Moreover, while the previous best result for different instances has been computed by a variety of solvers, our algorithm can now be used as a single tool to compute the result.

\vfill
\section{Preliminaries}
\label{preliminaries}
\subsection{Basic Concepts}
Let $G=(V=\{0,\ldots, n-1\},E)$ be an undirected graph and $N(v)\Is
  \setGilt{u}{\set{v,u}\in E}$ denote the neighbors of $v$.  The
  degree of a vertex $v$ is $d(v):=|N(v)|$.        The problem that we tackle in this paper is the \emph{graph clustering} problem.
    A clustering $\mathcal{C}$ is a partition of the set of vertices, \ie  a set of disjoint \emph{blocks/clusters} of vertices
  $V_1$,\ldots,$V_k$ such that $V_1\cup\cdots\cup
  V_k=V$. However, $k$ is usually not given in advance.
    A size-constrained clustering constrains the size of the blocks of a clustering by a given upper bound~$U$.
A clustering is \emph{trivial} if there is only one block, or all clusters/blocks contain only one element, i.e., are singletons. We identify a cluster $V_i$ with its node-induced subgraph of $G$. 
    The set $E(\mathcal{C}) := E \cap (\cup_i V_i \times V_i)$ is the set of \emph{intra-cluster edges}, and $E \setminus E(\mathcal{C})$ is the set of \emph{inter-cluster edges}. 
We set $|E(\mathcal{C})| =: m(\mathcal{C})$ and $|E \backslash E(\mathcal{C})| =: \overline{m}(\mathcal{C})$.
An edge running between two blocks is called \emph{cut edge}. There are different objective functions that are optimized in the literature. We review some of them in Section~\ref{subsec:objectives}. Our main focus in this work is on \emph{modularity}. 
However, our algorithm can be generalized to optimize other objective functions.
                            The \emph{graph partitioning problem} is also looking for a partition of the vertices. 
                           Here, a \emph{balancing constraint} demands that all blocks
have weight $|V_i|\leq (1+\epsilon)\lceil\frac{|V|}{k}\rceil=:L_{\max}$
for some imbalance parameter
$\epsilon$.      
 A vertex is a \emph{boundary vertex} if it is incident to a vertex in a different~block. 
The objective is to minimize the total
  \emph{cut} $\omega(E\cap\bigcup_{i<j}V_i\times V_j)$.

    In general, our input graphs $G$ have unit edge weights and vertex weights.
    However, even those will be translated into weighted problems in the course of a multi-level algorithm.
    In order to avoid tedious notation, $G$ will denote the current state of the graph before and after a (un)contraction in the multi-level scheme
    throughout this paper.
    \vspace*{-.25cm}
\subsection{Ensemble/Overlay Clusterings}
In machine learning, ensemble methods combine multiple weak classification algorithms to obtain a strong classifier.
These base clusterings are used to decide whether pairs of vertices 
should belong to the same cluster~\cite{OvelgoenneG13ensemble,StaudtM15engineering}. 
Given two clusterings, the \emph{overlay clustering} is a clustering in which two vertices belong to the same cluster if and only if they belong to the same cluster in each of the input clusterings. 
More precisely, given two clusterings $\mathcal{C}_1$ and $\mathcal{C}_2$ the \emph{overlay clustering} is the clustering where each block corresponds to a connected component of the graph $G_\mathcal{E} = (V,E\backslash \mathcal{E})$ where $\mathcal{E}$ is the union of the cut edges of $\mathcal{C}_1$ and $\mathcal{C}_2$, i.e. all edges that run between blocks in either $\mathcal{C}_1$ or $\mathcal{C}_2$.
Intuitively, if the input clusters agree that two vertices belong to the same cluster, the two vertices belong together with high confidence.
It is easy to see that the number of clusters in the overlay clustering cannot
be smaller than the number of clusters in each of~the~input~clusterings. 

An overlay clustering can be computed in expected linear-time.
More precisely, given two clusterings $\{\mathcal{C}_1, \mathcal{C}_2\}$, we use the following approach to compute the overlay clustering. 
Initially, the overlay clustering $\mathcal{O}$ is set to the clustering $\mathcal{C}_1$. 
We then iterate through the remaining clusterings and incrementally update the current solution $\mathcal{O}$. 
To this end, we use pairs of cluster IDs $(i,j)$ as a key in a hash map $\mathcal{H}$, where $i$ is a cluster ID of $\mathcal{O}$ and $j$ is a cluster ID of the current clustering $\mathcal{C}$. 
We initialize $\mathcal{H}$ to be the empty hashmap and set a counter $c$ to zero.
Then we iterate through the nodes. 
Let $v$ be the current node. 
If the pair $(\mathcal{O}[v],\mathcal{C}[v])$ is not contained in $\mathcal{H}$, we set $\mathcal{H}(\mathcal{O}[v],\mathcal{C}[v])$ to $c$ and increment $c$ by one.
Afterwards, we update the cluster ID of $v$ in $\mathcal{O}$ to be $\mathcal{H}(\mathcal{O}[v],\mathcal{C}[v])$.  
At the end of the algorithm, $c$ is equal to the number of clusters contained in the overlay clustering and each vertex is labeled with its cluster ID in $\mathcal{O}$.

\subsection{Objective Functions}
\label{subsec:objectives}
There are a variety of measures used to assess the quality of a clustering, such as \emph{coverage}~\cite{brandes2005network}, \emph{performance} \cite{van2001graph}, \emph{inter-cluster conductance}~\cite{kannan2004clusterings}, \emph{surprise}~\cite{arnau2004iterative}, \emph{map equation}~\cite{rosvall2009map} and \emph{modularity}~\cite{newman2004finding}. 
The most simple index realizing a traditional measure of clustering quality
is coverage. The coverage of a graph clustering $\mathcal{C}$ is defined as the fraction of intra-cluster edges  within the complete set of edges
$\text{cov}(\mathcal{C}) := \frac{m(\mathcal{C})}{m}  $.
Intuitively, large values of coverage correspond to a good quality of a clustering. However,
one principal drawback of coverage is that the converse is not necessarily true: coverage
takes its largest value of 1 in the trivial case where there is only one cluster. 
\emph{Modularity} fixes this issue by comparing the coverage of a clustering to the same value after rearranging edges randomly.
\emph{Performance} is the fraction of correctly classified vertex pairs, w.r.t the set of edges. 
\emph{Inter-cluster conductance} returns the worst (i.e. the thickest) bottleneck created by separating a cluster from the rest of the graph. \emph{Surprise} measures the probability that a random graph $\mathcal{R}$ has more intra-cluster edges.
\emph{Map equation} is a flow-based  and  information-theoretic method to assess clustering quality.
Our focus in this work is on modularity as it is a widely accepted quality function and has been the main objective function in the 10th DIMACS implementation challenge~\cite{dimacschallengegraphpartandcluster}. For further discussions of these indices we refer the reader to the given references, and simply state the formal definition of modularity here: \[
\mathcal{Q}(\mathcal{C})  := \cov(\mathcal{C})- \mathbb{E}[\cov(\mathcal{C})]
                         =\frac{m(\mathcal{C})}{m}- \frac{1}{4m^2} \sum_{V_i \in \mathcal{C}}\left( \sum_{v \in V_i} d(v) \right)^2
\]

\vspace*{-.5cm}
\subsection{Related Work}
This paper is a summary and extension of the bachelor's thesis by Sonja Biedermann~\cite{baBiedermann}.
There has been a \emph{significant} amount of research on graph clustering. We refer the reader to \cite{fortunato2010community,hartmann2016clustering} for thorough reviews of the results in this area. 
Here, we focus on results closely related to our main contributions. 
It is common knowledge that there is no single best strategy for graph clustering. Moreover, most quality indices for graph clusterings have turned out to be NP-hard to optimize and rather resilient to effective approximations, see e.g. \cite{ausiello2012complexity,brandes2008modularity,wagner1993between}, allowing only heuristic approaches for optimization. 
Other approaches often rely on specific strategies with high running times, e.g. the iterative removal of central edges \cite{newman2004finding}, or the direct identification of dense subgraphs \cite{derenyi2005clique}. Provably good methods with a decent running time include algorithms that have a spectral embedding of the vertices as the basis for a geometric clustering~\cite{brandes2007engineering}, min-cut tree clustering~\cite{flake2004graph}, a technique which guarantees certain bounds on bottlenecks and an approach which relies on random walks in graphs staying inside dense regions with high probabilities~\cite{van2001graph}. 
The Louvain method is a multi-level clustering algorithm introduced by Blondel~\etal\cite{louvain} that optimizes modularity as an objective function. 
As we use this method in our algorithm to create the initial population as well as to improve individuals after recombination, we go into more detail for that method in Section~\ref{method:louvain}.

Recently, the 10th DIMACS challenge on graph partitioning and graph clustering compared different state-of-the-art graph clustering algorithms w.r.t optimizing modularity~\cite{benchmarksfornetworksanalysis}. Most of the best results have been obtained by Ovelgönne and Geyer-Schulz \cite{OvelgoenneG13ensemble} as well as Aloise \etal\cite{11}. CGGC \cite{OvelgoenneG13ensemble} used an ensemble learning strategy during the challenge. From several weak clusterings an overlap clustering is computed and a more expensive clustering algorithm is applied. VNS \cite{11} applies the Variable Neighborhood Search heuristic to the graph clustering problem. 
Later, D{\v{z}}ami{\'{c}} \etal\cite{Dzamic2017} outperformed VNS by proposing an ascent-decent VNS. 
We compare ourselves to the results above in~Section~\ref{evaluation}.

\emph{Evolutionary Graph Clustering.} Tasgin and Bingol~\cite{tasgin} introduce a genetic algorithm using modularity as
a quality measure. They chose an integer encoding for representing the
population clusterings. Individuals are randomly initialized, with some bias
for assigning direct neighbors to the same cluster. Recombination is done one way, \ie instead of using mutual exchange, clusters are transferred from a source individual to a target individual.
More precisely, the operation picks a random vertex in the source individual and clusters all vertices of its cluster together in the destination individual. This is repeated several times in one recombination operation. 
As for mutation, the authors chose to pick one
vertex and move it to a random cluster, which is almost guaranteed to decrease
the fitness. This approach does not use local search to improve individuals.

A mutation-less agglomerative hierarchical genetic algorithm is presented by Lipczak and
Milios in~\cite{aga}.  The authors represent each cluster as one individual and use one-point and uniform crossovers in that representation. 
Two synthetic networks have been used for the experimental evaluation. An important advantage of this algorithm is the possibility to distribute its computations.

\subsection{Karlsruhe High-Quality Partitioning}
\label{s:kaHIP}
Within this work, we use the open source multi-level graph partitioning framework KaHIP~\cite{kabapeE} (Karlsruhe High-Quality Partitioning).
More precisely, we employ partitioning tools contained therein to create high-quality partitions of the graphs. 
We shortly outline its main components. 
KaHIP implements many different algorithms, for example, flow-based methods and more-localized local searches within a multi-level framework, as well as several coarse-grained parallel and sequential meta-heuristics. 
Recently, specialized methods to partition social networks and web graphs have been included in the framework \cite{sclp}. 

\subsection{Multi-level Louvain Method}
\label{method:louvain}
The Louvain method is a multi-level clustering algorithm introduced by Blondel~\etal\cite{louvain}. 
It is an approach to graph clustering that optimizes modularity as an objective function. Since we instantiate our memetic algorithm to optimize for modularity, we give more detail in order to be self-contained.
The algorithm is organized in two phases: a local movement phase and contraction/uncontraction phase. 
The first phase, \emph{local movement}, works in rounds and is done as follows: In the beginning, each vertex is a singleton cluster. Vertices are then traversed in random order and always moved to the neighboring cluster yielding the highest modularity increase. Computing the best move can be done in time proportional to the degree of a vertex by storing cumulative vertex degrees of clusters. More precisely, the gain in modularity by removing a vertex from a cluster can be computed by \[
     \Delta Q(u) = \frac{s}{m} + \frac{d(u)}{4m^2} + \frac{(\sum_{v \in C} d(v))^2}{4m^2} - \frac{(d(u) + \sum_{v \in C} d(v))^2}{4m^2}.\]
A similar formula holds when adding a singleton vertex to a cluster. Once a vertex is moved, the cumulated degrees of the affected clusters are updated. Hence, the best move can be found in time proportional to the degree of a vertex. The local movement algorithm stops when a local maximum of the modularity is attained, that is when no vertex move a yields modularity gain. 
The \emph{second phase} of the algorithm consists in contracting the clustering. Contracting the clustering works as follows: 
each block of the clustering is contracted into a single node. 
There is an edge between two vertices $u$ and $v$ in the contracted graph if the
two corresponding blocks in the clustering are adjacent to each other in $G$,
\ie block $u$ and block $v$ are connected by at least one edge.
The weight of an edge $(A,B)$ is set to the sum of the weight of the edges that run between block $A$ and block $B$ of the clustering. Moreover, a self-loop is inserted for each vertex in the contracted graph. The weight of this edge is set to be the cumulative weight of the edges in the respective cluster.
Note that due to the way the contraction is defined, a clustering of the coarse graph corresponds to a clustering of the finer graph with the same objective. 
The algorithm then continues with the local movement phase on the contracted graph.
In the end, the clustering contractions are undone and at each level local movement improves the current clustering w.r.t. modularity.

\section{Memetic Graph Clustering}
\label{memeticalgorithmforgraphclustering}
We now explain the components of our memetic graph clustering algorithm. 
Our algorithm starts with a population of individuals (in our case one individual is a clustering of the graph) and evolves the population into different populations over several rounds. 
In each round, the EA uses a selection rule based on the fitness of the individuals (in our case the objective function of the clustering problem under consideration) of the population to select good individuals and recombine them to obtain improved offspring. 
Our selection process is based on the tournament selection rule \cite{Miller95geneticalgorithms}, i.e. $\mathcal{C}$ is the fittest out of two random individuals $R_1, R_2$ from the population. 
When an offspring is generated an elimination rule is used to select a member of the population and replace it with the new offspring. 
In general, one has to take both into consideration, the fitness of an individual and the distance between individuals in the population~\cite{DBLP:conf/gecco/PorumbelHK11} in order to avoid premature convergence of the algorithm. 
We evict the solution that is \emph{most similar} with respect to the edges that run in between clusters with the offspring among those individuals in the population that have a worse or equal objective than the offspring itself. If there is no such individual, then the offspring is rejected and not inserted into the population.
The difference between two individuals is defined as the size of the symmetric difference between their sets of cut edges. 
Our algorithm generates only one~offspring~per~generation. 

The core of our algorithm are our novel recombination and mutation operations. We provide two different kinds of operations, flat- and multi-level recombination operations.
We also define a mutation operator that splits clusters by employing graph partitioning techniques. In any case, the offspring is typically improved using a local search algorithm that optimizes for the objective function of the graph clustering problem. We continue this section by explaining the details of our algorithm with a focus on modularity---the results are transferable to other clustering problems by using local search algorithms that optimize the objective function of the problem under consideration.

\subsection{Initialization/Creating Individuals}
\label{ss:init}

We initialize our population using the following modified Louvain algorithm. We also use the following algorithm to create an individual in recombine and mutation operations. Depending on the objective of the evolutionary algorithm, the choice of algorithms to initialize the population may be different. 
We describe our scheme for modularity.

We mainly use the Louvain
method to create clusterings. Due to the non-deterministic nature of this algorithm---the order of nodes
visited is randomized---we obtain different initial graph clusterings and later on individuals.
In order to introduce more diversification, we modify the approach by using a different coarsening strategy based on the size-constrained label propagation algorithm~\cite{sclp}.
Label propagation works similar to the local movement phase of the Louvain method. However, the objective of the algorithm is different.
Without size-constraints, it was proposed by Raghavan~\etal\cite{labelpropagationclustering}.
Initially, each vertex is in its own cluster/block, \ie the initial block ID of a vertex is set to its vertex ID.
The algorithm then works in rounds. 
In each round, the vertices of the graph are traversed in a random order. 
When a vertex $v$ is visited, it is \emph{moved} to the block that has the strongest connection to $v$, \ie it is moved to the cluster $V_i$ that maximizes $\omega(\{(v, u) \mid u \in N(v) \cap V_i \})$ whilst not overloading the target cluster w.r.t to the size-constraint bound $U$. 
Ties are broken randomly. 
The process is repeated until the process has converged. 
Here, we perform at most $\ell$ rounds of the algorithm, where $\ell$ is a tuning parameter, and stop the algorithm if less than five percent of the vertices changed its cluster during one round.
One LPA round can be implemented to run in $\Oh{n+m}$ time.

We modify the coarsening stage of the Louvain method using size-constrained label propagation as follows: 
For the first $\lambda\in_{\text{rnd}} [0,4]$ levels of the multi-level hierarchy we use size-constraint label propagation to compute the clustering to be contracted instead of the local movement phase to compute a clustering. We choose $U \in_{\text{rnd}} [n/10,n]$ in the beginning of the multi-level algorithm. Afterwards, we switch to the local movement phase of the Louvain method to compute a clustering to be contracted. Note that for $\lambda = 0$ the method is the Louvain method. In any case, in the uncoarsening phase local movement improves the current clustering by optimizing modularity.

\subsection{Recombination}

We define flat- and multi-level recombination operations which we are going to explain now.
\emph{Both} operations use the notion of overlay clustering to pass on good parts of the solutions to the offspring.
Some of our recombination operators ensure that the offspring has non-decreasing fitness. Moreover, our recombine operations can recombine a solution from the population with an arbitrary clustering~of~the~graph.
Due to the fact that our local search and multi-level algorithms are randomized, a recombine operation performed twice using the same parents can yield different offsprings. 
\subsubsection*{Flat Recombination}
The  \emph{basic flat recombination operation} starts by taking two clusterings $\mathcal{C}_1, \mathcal{C}_2$ from the population as input and computes its overlay. The overlay is then contracted such that 
    vertices represent blocks in the overlay clustering and edges represent edges between vertices of two blocks.
    Edge weights are equal to the summed weight of the edges that run between the respective blocks, and self-loops are inserted with the edge weight being equal to the summed internal edge weight of that corresponding block.
    Note that due to the way contraction is defined, a clustering of the contracted graph can be transformed into a clustering of the input network having the same modularity score. 
Hence, we use the Louvain method to cluster the contracted graph, which then constitutes the offspring.
Also, note that the contraction of the overlay ensures that vertices that are clustered together in both inputs will belong to the same cluster in the offspring. Only vertices which are split by one of the input clusterings will be affected by the Louvain method. What follows are multiple variations of the basic flat recombination method that ensure non-decreased fitness of the input individual, \ie the offspring has fitness at least as good as the better input individual.

\emph{Apply Input Clustering.}
This operator uses the better of the two parent clusterings as a starting point for the Louvain method on the contracted graph.
Due to the way contraction is defined the objective function of the starting point on the contracted graph is the same as of the corresponding clustering on the input network.
The resulting offspring can be expected to be
less diverse than when using the plain recombination operator, but may significantly more improve upon the parent. 

\emph{Cluster-/Partition Recombination.}
This operator differs from the previous two operators insofar that only one of the
parents is selected from the population. The other parent is manufactured on the
spot by running either the size-constrained label propagation (SCLP) or a graph partitioning algorithm on the
input graph. The resulting clustering can be very different from to the other parent. This introduces more diversification which helps local search to explore a larger search space. 
When using SCLP clustering for the recombination, we use the parameters as described in Section~\ref{ss:init}.
When using a partition, we use the KaHIP graph partitioning framework with the fastsocial preconfiguration using $k \in_\text{rnd} [2, 64]$ and $\epsilon \in_\text{rnd} [0.03, 0.5]$. 

\subsubsection*{Multi-level Recombination}
We now explain our multi-level recombine operator.
This recombine operator also ensures that the solution quality of the offspring is \emph{at least as good as the best of both parents}.
For our recombine operator, let $\mathcal{C}_1$ and $\mathcal{C}_2$ again be two individuals from the population. 
Both individuals are used as input for the multi-level Louvain method in the following sense. 
Let $\mathcal{E}$ be the set of edges that are cut edges, \ie edges that run between two blocks, in either $\mathcal{C}_1$ \emph{or} $\mathcal{C}_2$. 
All edges in $\mathcal{E}$ are blocked during the coarsening phase, \ie they are \emph{not} contracted during the coarsening phase.
In other words, these edges cannot be contracted during the multi-level scheme. 
We ensure this by modifying the local movement phase of the Louvain method, \ie clusters can only grow inside connected components of the overlay $(V,E\backslash \mathcal{E})$.
We stop contracting clusterings when no contractable edge is left. 

When coarsening is stopped, we then apply the better out of both input individuals w.r.t. the objective to the coarsest graph and use this as initial clustering instead of running the Louvain method on the coarsest graph. However, during uncoarsening local search still optimizes modularity on each level of the hierarchy.
Note that this is possible since we did not contract any cut edge of both inputs.
Also, note that this way we obtain a clustering of the coarse graph having a modularity score being equal to the better of both input individuals. 
Local movement guarantees no worsening of the clustering. Hence, the offspring is at least as good as the best input individual.
Note that the coarsest graph is the same as the graph obtained in a flat recombination. However, the operation now is able to pass on good parts of the solution on multi-levels of the hierarchy during uncoarsening and hence has a more fine-grained view on both individuals.

\subsection{Mutation}
Our recombination operators can only decrease the number of clusters present in the solution. To counteract this behavior, we define a mutation operator that selects a subset of clusters and splits each of them in half. Splitting is done using the KaHIP graph partitioning framework. The number of clusters to be split is set to $p_s|\mathcal{C}|$ where $p_s$ is the splitting probability and $|\mathcal{C}|$ is the number of clusters that the clustering selected from the population has. 
That means that we split a cluster into two balanced blocks such that there are only a small amount of edges running between them.
We do this operation with two individuals from the population that are found by tournament selection. 
This results in two output individuals which are used as input to a multi-level recombination operation.
The computed offspring of that operation is then inserted into the population as described above.
\vfill
\subsection{Parallelization}
\label{s:parallelization}
We now explain the island-based parallelization that we use. We use a parallelization scheme that has been successfully used in graph partitioning~\cite{kaffpaE}. Each processing element (PE)  basically performs the same operations using different random seeds.
First, we estimate the population size~$\mathcal{S}$: each PE creates an individual and measures the time $\overline{t}$ spent. 
We then choose $\mathcal{S}$ such that the time for creating $\mathcal{S}$ clusterings is approximately $t_{\text{total}}/c$ where the fraction $c$ is a tuning parameter and $t_{\text{total}}$ is the total running time that the algorithm is given to produce a clustering of the graph. 
The minimum amount of individuals in the population is set to~3, the maximum amount of the individuals in the population is set to 100. 
The lower bound on the population size is chosen to ensure a certain minimum of diversity, while
the upper bound is used to ensure convergence. 

Each PE then builds its own population. 
Afterwards, the algorithm proceeds in rounds as long as time is left. 
Either a mutation or recombination operation is performed. 
Our communication protocol is similar to \textit{randomized rumor spreading} which has shown to be scalable in previous work~\cite{kaffpaE}. 
Let $p$ denote the number of PEs used. A communication step is organized in rounds. 
In each round, a PE chooses a communication partner and sends her the currently best individual $\mathcal{C}$ of the local population. 
The selection of the communication partner is done uniformly at random among those PEs to which $\mathcal{C}$ not already has been sent to.  
Afterwards, a PE checks if there are incoming individuals and if so inserts them into the local population using the elimination strategy described above.
If the best local individual has changed, all PEs are again eligible.
This is repeated $\log p$ times.
The algorithm is implemented \textit{completely asynchronously}, i.e. there is no need for a global synchronization.

\section{Experimental Evaluation}
\label{evaluation}
\subsubsection*{System and Methodology}\label{Methodology}
We implemented the memetic algorithm described in the previous section within the KaHIP (Karlsruhe High Quality Partitioning) framework. 
The code is written in C++ and MPI. It has been compiled using g++-5.2 with flags \texttt{-O3} and OpenMPI 1.6.5.
We refer to the algorithm presented in this paper as VieClus. 
We plan to further improve the code and then to release it to make it available to a larger audience.
Throughout this section, our main objective is modularity.
All experiments comparing VieClus with competing algorithms are performed on a cluster with $512$ nodes, where each node has two Intel Xeon E5-2670 Octa-Core (Sandy Bridge) processors 
clocked at $2.6$ GHz, $64$~GB main memory, $20$ MB L3- and 8x256 KB L2-Cache and runs RHEL 7.4.
We use the \emph{arithmetic mean} when averaging over solutions of the same instance and the \emph{geometric mean} when averaging over different instances
in order to give every instance a comparable influence on the final result.
It is well known that the algorithms that scored most of the points during the 10th DIMACS challenge compute better results than the Louvain method. 
Moreover, LaSalle~\cite{lasalle2015graph} reports results for his Nerstrand algorithm that are on average equal or slightly better than the Louvain method on the instances used here, but the best results of this method are consistently worse than the result computed during the DIMACS challenge.
Hence, we refrain from doing additional experiments~with~the~Louvain~method. 

\vfill
\begin{table}[tb]   
\begin{center}
\begin{footnotesize}
\caption{Results of our algorithm on the benchmark test set. Columns from left to right: average modularity achieved by our algorithm, $\overline{t}$ average time in minutes needed to beat the old challenge result, the best score computed of our algorithm, the best modularity scores achieved by challenge participants, running time in minutes needed to create the previous best entry according to \cite{lasalle2015graph} and reference to the solver that achieved the result during the 10th DIMACS challenge. Entries that improve or reproduce the result of the implementation challenge are highlighted.}
\label{fig:modularityscores}
\label{tab:test_instances}
\begin{tabular}{|l||l|r|l|||l|r|r|}
\hline
Graph & Avg. $\mathcal{Q}$& $\overline{t}$ [m]& Max. $\mathcal{Q}$& $\mathcal{Q}$~\cite{dimacschallengegraphpartandcluster}  & $t_\text{sol}$[m]& Solver \\
\hline
\hline
\multicolumn{7}{|c|}{Small Instances} \\
\hline
\hline
\texttt{as-22july06}                   & \textbf{\numprint{0.679391}} & <1               & \textbf{\numprint{0.679396}} & \numprint{0.678267} & \numprint{6.6}   & CGGC\cite{OvelgoenneG13ensemble}\\
\texttt{astro-ph}                      & \textbf{\numprint{0.746285}} & <1               & \textbf{\numprint{0.746292}} & \numprint{0.744621} & \numprint{11.9}  & VNS\cite{11}\\
\texttt{celegans\_metabol}             & \textit{\numprint{0.453248}} & <1               & \textit{\numprint{0.453248}} & \numprint{0.453248} & <1               & VNS\cite{11}\\
\texttt{cond-mat-2005}                 & \textbf{\numprint{0.750065}} & <1               & \textbf{\numprint{0.750171}} & \numprint{0.746254} & \numprint{40.9}  & CGGC\cite{OvelgoenneG13ensemble}\\
\texttt{email}                         & \textit{\numprint{0.582829}} & <1               & \textit{\numprint{0.582829}} & \numprint{0.582829} & <1               & VNS\cite{11}\\
\texttt{PGPgiantcompo}                 & \textbf{\numprint{0.886853}} & <1               & \textbf{\numprint{0.886853}} & \numprint{0.886564} & \numprint{1.9}     & CGGC\cite{OvelgoenneG13ensemble}\\
\texttt{polblogs}                      & \textit{\numprint{0.427105}} & <1               & \textit{\numprint{0.427105}} & \numprint{0.427105} & <1     & VNS\cite{11}\\
\texttt{power}                         & \textbf{\numprint{0.940975}} & <1               & \textbf{\numprint{0.940977}} & \numprint{0.940851} & <1     & VNS\cite{11}\\
\texttt{smallworld}                    & \textbf{\numprint{0.793186}} & <1               & \textbf{\numprint{0.793187}} & \numprint{0.793042} & \numprint{16.8}     & VNS\cite{11}\\
\texttt{memplus}                       & \textbf{\numprint{0.701242}} & \numprint{2.6}   & \textbf{\numprint{0.701275}} & \numprint{0.700473} & \numprint{3.2}     & CGGC\cite{OvelgoenneG13ensemble}\\
\texttt{G\_n\_pin\_pout}               & \textbf{\numprint{0.500457}} & \numprint{3.9}   & \textbf{\numprint{0.500466}} & \numprint{0.500098} & \numprint{64,8}     & CGGC\cite{OvelgoenneG13ensemble}\\
\texttt{caidaRouterLevel}              & \textbf{\numprint{0.872804}} & \numprint{5.0}   & \textbf{\numprint{0.872828}} & \numprint{0.872042} & \numprint{81.0}     & CGGC\cite{OvelgoenneG13ensemble}\\
\texttt{rgg\_n17}                      & \textbf{\numprint{0.978448}} & \numprint{5.0}   & \textbf{\numprint{0.978454}} & \numprint{0.978324} & \numprint{37,5}     & VNS\cite{11}\\
\texttt{luxembourg.osm}                & \textbf{\numprint{0.989665}} & \numprint{7.3}   & \textbf{\numprint{0.989672}} & \numprint{0.989621} & \numprint{40,9}     & VNS\cite{11}\\
\hline
\hline
\multicolumn{7}{|c|}{Large Instances}\\
\hline
\hline
\texttt{coAuthorsCiteseer}             & \textbf{\numprint{0.906804}} & \numprint{3.9}   & \textbf{\numprint{0.906830}} & \numprint{0.905297} & \numprint{91,3}     & CGGC\cite{OvelgoenneG13ensemble}\\
\texttt{citationCiteseer}              & \textbf{\numprint{0.825518}} & \numprint{12.9}  & \textbf{\numprint{0.825545}} & \numprint{0.823930} & \numprint{77,6}     & CGGC\cite{OvelgoenneG13ensemble}\\
\texttt{coPapersDBLP}                  & \textbf{\numprint{0.868019}} & \numprint{20.5}  & \textbf{\numprint{0.868058}} & \numprint{0.866794} & \numprint{603.3}     & CGGC\cite{OvelgoenneG13ensemble}\\
\texttt{belgium.osm}                   & \textbf{\numprint{0.995062}} & \numprint{29.5}  & \textbf{\numprint{0.995064}} & \numprint{0.994940} & \numprint{102,9}     & CGGC\cite{OvelgoenneG13ensemble}\\
\texttt{ldoor}                         & \textbf{\numprint{0.970521}} & \numprint{35.1}  & \textbf{\numprint{0.970555}} & \numprint{0.969370} & \numprint{485,6}     & ParMod\cite{10}\\
\texttt{eu-2005}                       & \textbf{\numprint{0.941575}} & \numprint{65.8}  & \textbf{\numprint{0.941575}} & \numprint{0.941554} & \numprint{341,5}     & CGGC\cite{OvelgoenneG13ensemble}\\
\texttt{in-2004}                       & \textbf{\numprint{0.980684}} & \numprint{237.4} & \textbf{\numprint{0.980690}} & \numprint{0.980622} & \numprint{244,0}     & CGGC\cite{OvelgoenneG13ensemble}\\
\texttt{333SP}                         & \textbf{\numprint{0.989316}} & \numprint{297.1} & \textbf{\numprint{0.989356}} & \numprint{0.989095} & \numprint{976,9}     & ParMod\cite{10}\\
\texttt{prefAttachment}                & \numprint{0.315843}          & *         & \textbf{\numprint{0.316089}} & \numprint{0.315994} & \numprint{1353,1}     & VNS\cite{11}\\
\hline
\end{tabular}
\end{footnotesize}
\end{center}
\vspace*{-.5cm}
\end{table}

\paragraph*{Parameters}
We \emph{did not} perform a tuning of the parameters of the algorithm, rather we chose the parameters described above and below to be reasonable and to introduce a large amount of diversification or we chose parameters that were a good choice in previous evolutionary algorithms~\cite{kaffpaE}.
As we our main design goal is to introduce as much diversification as possible, we use all recombination and mutation operations in our algorithm. 
The ratio of mutation to recombine operations has been set to 1:9 as this has been a good choice in previous evolutionary algorithms~\cite{kaffpaE}. 
When we perform a recombine operation, we pick the recombine operation uniformly at random and diversify the parameters as described above. 
When performing a mutation operation, we use a splitting probability $p_s$ uniformly at random in~$[0.01,0.1]$. We invest 1/10 of the total time to create the initial population. 
\paragraph*{Instances}
We use the graphs that have been used for the 10th DIMACS implementation challenge on graph clustering and graph partitioning~\cite{dimacschallengegraphpartandcluster}.
A list of the instances as well as the modularity scores that have been obtained during the challenge can be found in Table~\ref{tab:test_instances}.
We exclude the instances \texttt{cage15}, \texttt{audikw1}, \texttt{er-fact1.5-scale25},  \texttt{kron\_*}, \texttt{uk-2002}, \texttt{uk-2007-05} from the challenge testbed in our evaluation, because they are either too large to be feasible for an evolutionary algorithm or they do not contain a significant cluster structure as indicated by the reported modularity score~\cite{benchmarksfornetworksanalysis}.

\vfill
\subsection{Evolutionary Graph Clustering}
\begin{figure}[t!]
\centering
\includegraphics[width=6.75cm]{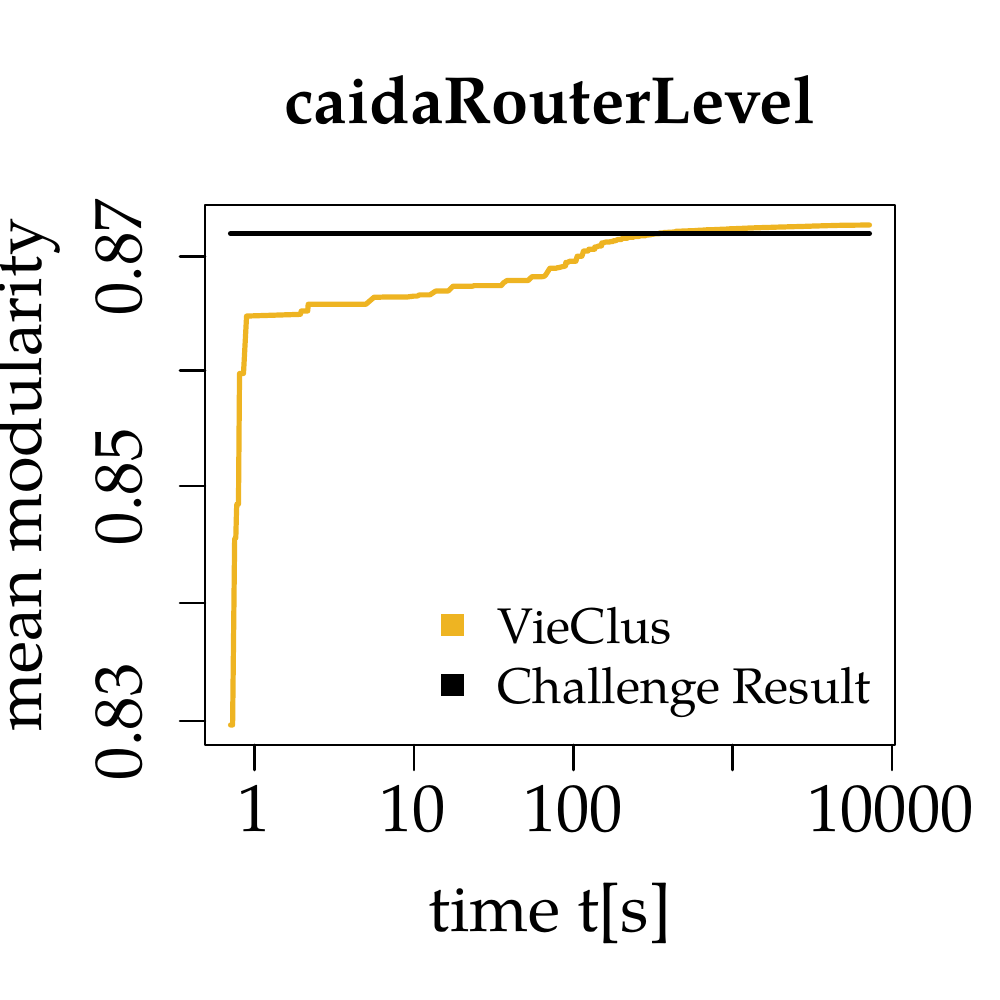} \includegraphics[width=6.75cm]{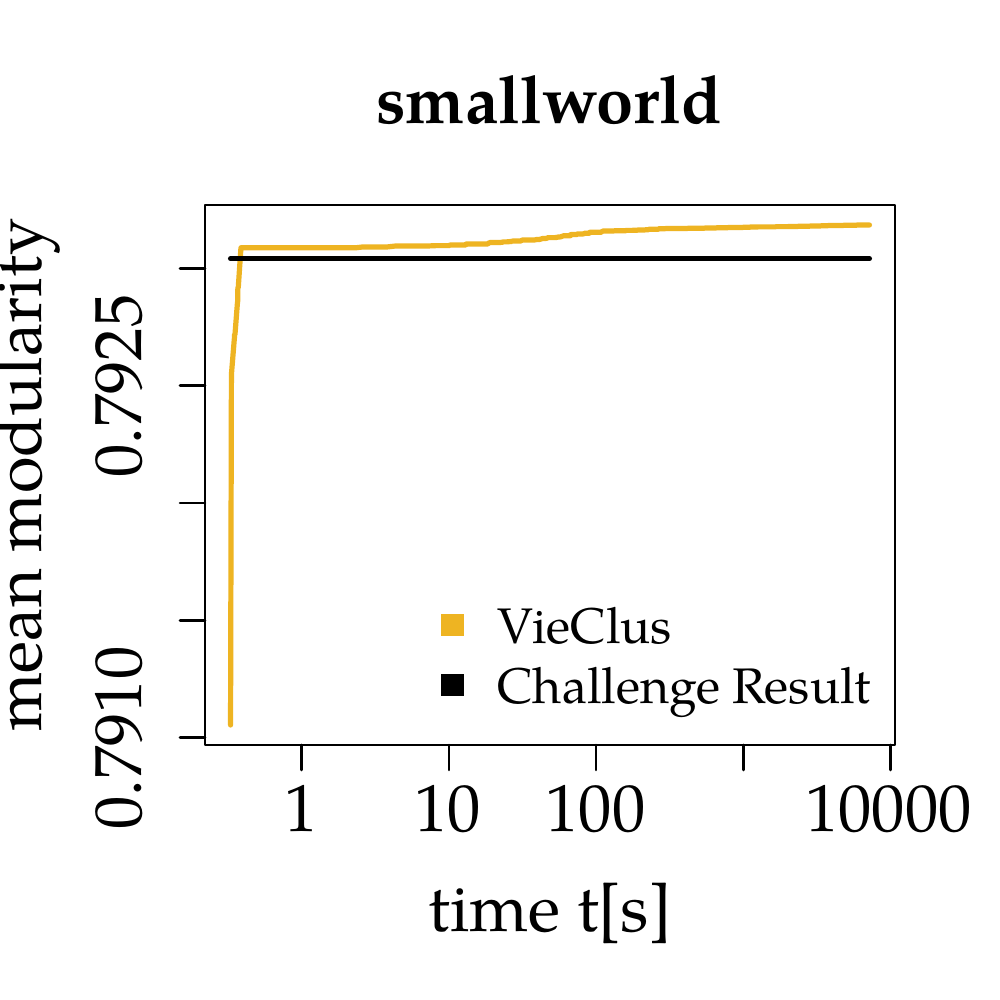}\\
        \vspace*{-.25cm}
\includegraphics[width=6.75cm]{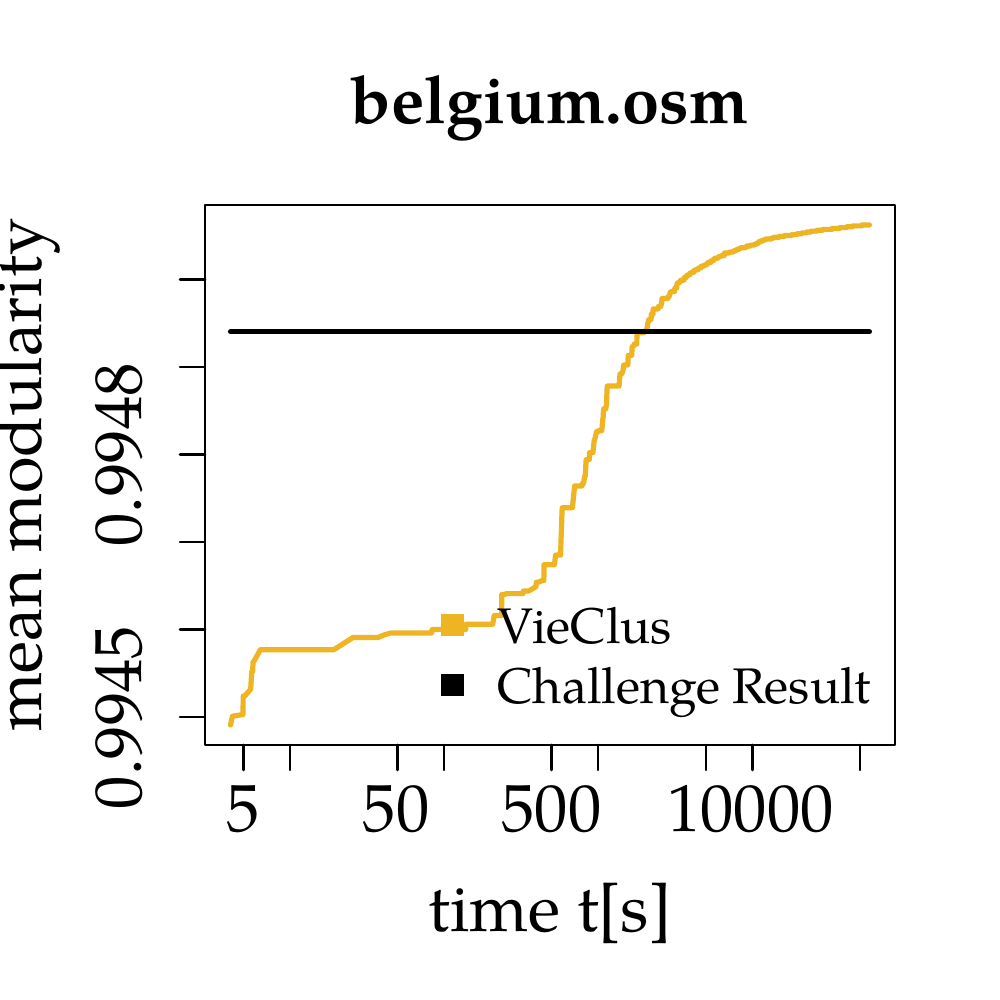} \includegraphics[width=6.75cm]{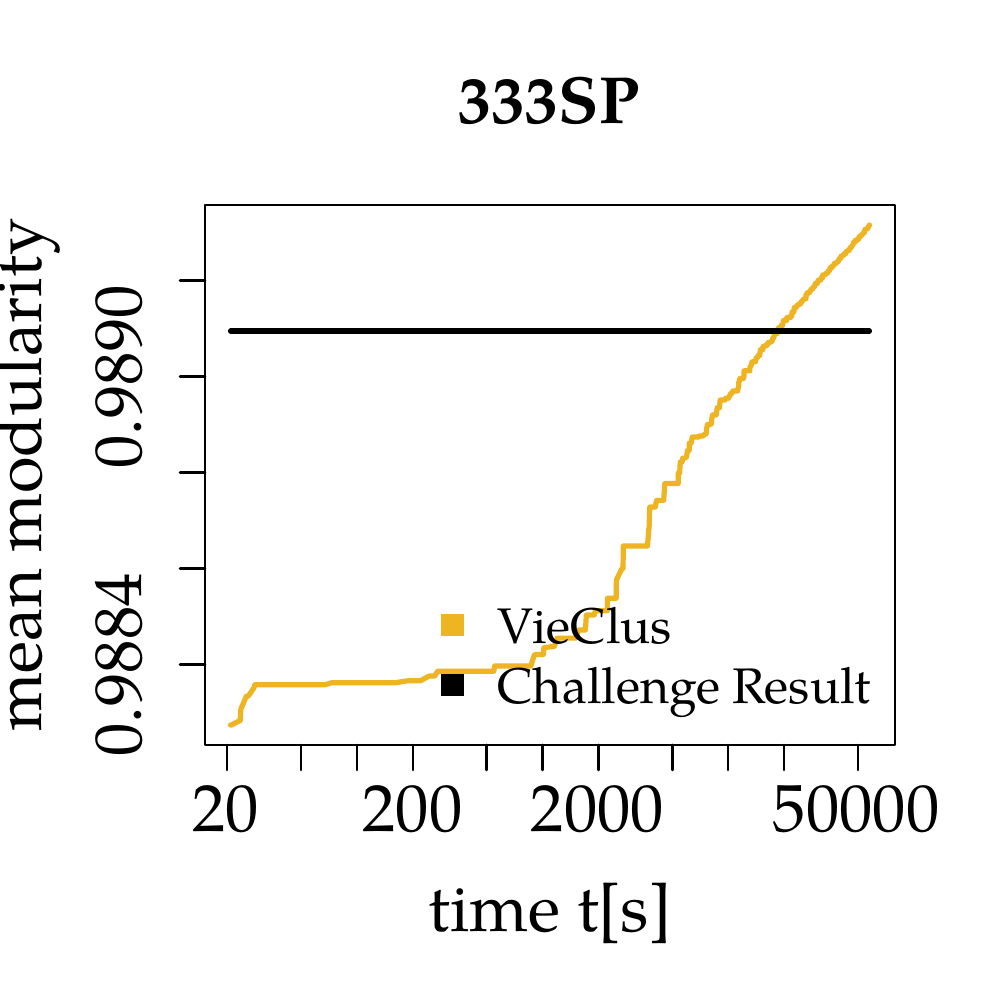}\\
        \vspace*{-.25cm}
\includegraphics[width=6.75cm]{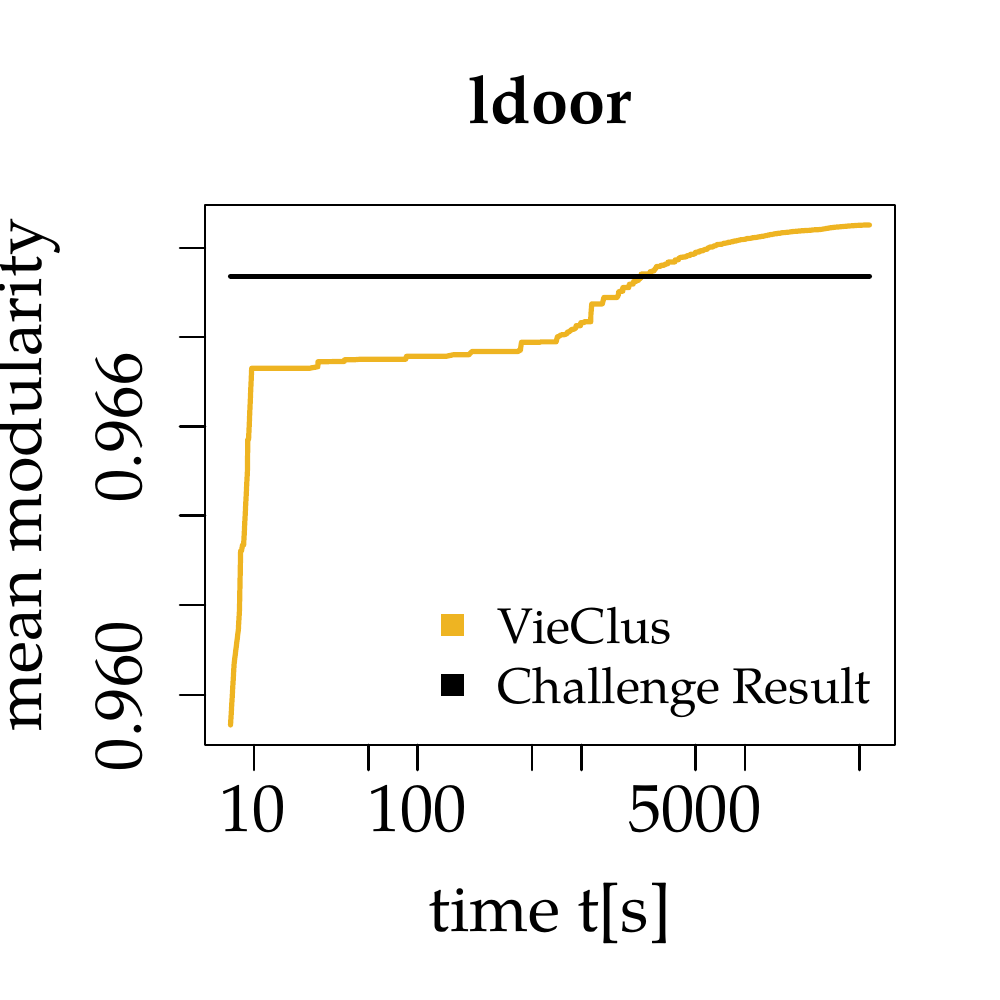} \includegraphics[width=6.75cm]{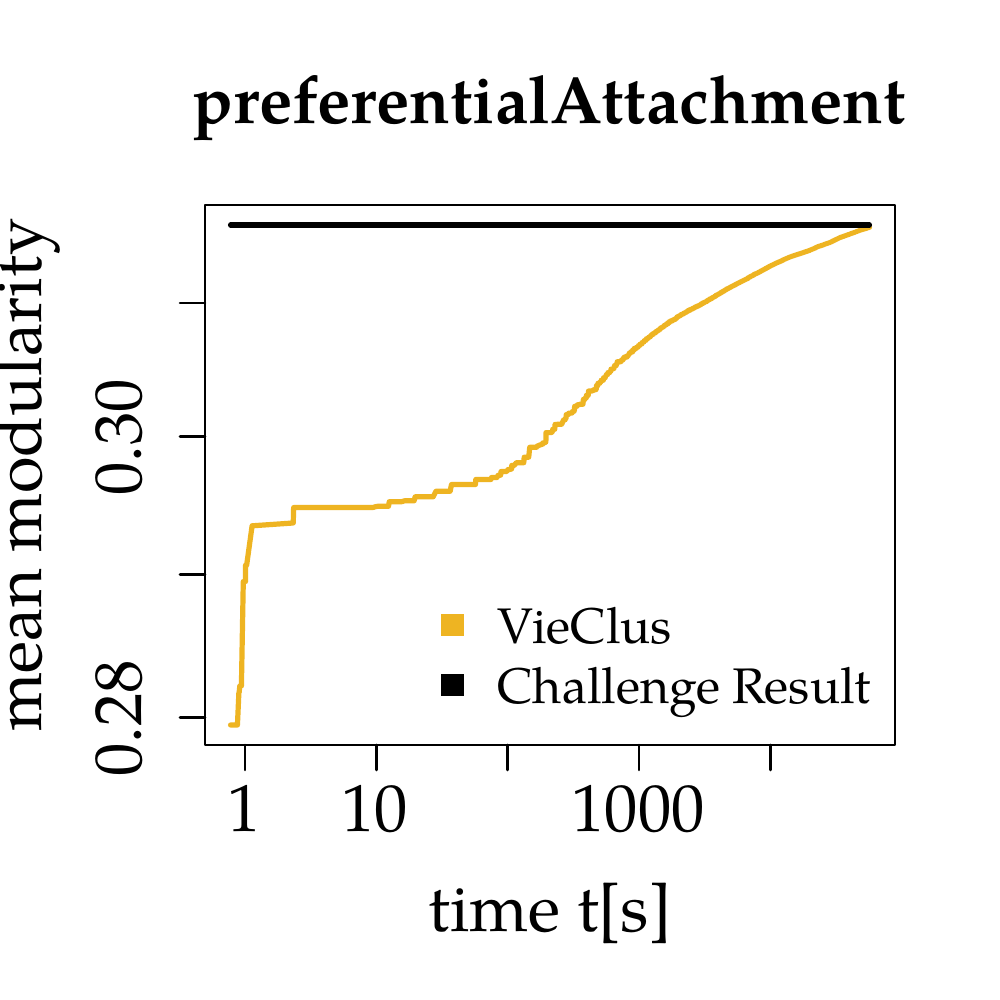}\\
        \vspace*{-.5cm}

\caption{Convergence plots for the instances the small instances \texttt{caidaRouterLevel}, \texttt{smallworld} and the large instances \texttt{333SP}, \texttt{belgium.osm}, \texttt{ldoor}, \texttt{preferentialAttachment}.}
\vspace*{1cm}
\label{fig:convergenceplots}
\end{figure}

We now run our algorithm on all the DIMACS instances under consideration using the rules used there, \ie
\emph{running time is not an issue} but we want to achieve modularity values as large as possible for each instance. 
On the small instances, we give our algorithm 2 hours of time to compute a solution and on the large instances, we set the time limit to 16 hours. In any case, we use 16 cores of our machine, \ie one node of the machine. We perform the test five times with different random seeds. Table~\ref{fig:modularityscores} summaries the results of our experiment, and Figure~\ref{fig:convergenceplots}  shows convergence plots for~a~selected~subset~of~the~instances.

First of all, on \emph{every} instance under consideration, our algorithm is able to compute a result that is better than the currently reported modularity value in literature. 
More precisely, in 98 out of 115 runs of our algorithm, the previous benchmark result of the 10th DIMACS implementation challenge is outperformed. In further 15 out of 115 runs, we reproduce the results. This is the case for each of the runs for the graphs \texttt{celegans\_metabolic}, \texttt{email} and \texttt{polblogs}. 
The two cases in which our algorithm does not beat the previous best solver, are 2 runs on the graph \texttt{prefAttachment}. The other 3 runs on that graph outperform the previous result.
The time needed to compute a clustering having a similar score on that graph is roughly 95\% of the total running time.
Moreover, as the convergence plots in~Figure~\ref{fig:convergenceplots} show this may be fixable by giving the algorithm a larger amount of time to compute~the~solution.
Overall, the time needed to outperform the previous benchmark results ranges from less than a minute to a couple of minutes on the small instances. 
On the large instances our algorithm needs more time, but on most instances, the previous benchmark result can be computed in roughly an hour. 
Table~\ref{fig:modularityscores} also reports the running time of the solver that obtained the result during the DIMACS challenge. 
Our algorithm is faster in every case compared to the previous solver (eventually by more than an order of magnitude), however, the machines used for the experiment are different and our algorithm is a parallel algorithm whereas previous solvers are sequential.
Note that the final improvements over the old result are fairly small (<0.1\% on average). This is not surprising, as previous solvers already invested a large amount of time to compute the results. However, note that the previous result has been computed by different solvers and our evolutionary algorithm can be seen as a single~tool~to~compute~the~result.

We now compare the results with recently published results~\cite{lasalle2015graph,hamann2017simple,lu2015parallel,ryu2016quick}.
LaSalle~\cite{lasalle2015graph} reports results on all graphs from the DIMACS challenge subset. However, each of the best computed result is worse than the result computed by the respective best algorithm during the DIMACS challenge (and hence worse compared to our algorithm).
Moreover, LaSalle~\cite{lasalle2015graph} reports that his Nerstrand algorithm is on average equal or slightly better than the Louvain method on the instances used here. 
Lu \etal \cite{lu2015parallel} present a result for \texttt{coPapersDBLP} ($\mathcal{Q} = \numprint{0.858088}$) and
Ryu and Kim \cite{ryu2016quick} report modularity for \texttt{email} ($\mathcal{Q} = \numprint{0.568}$) which are worse compared to the results that we report here.
Hamann~\etal~\cite{hamann2017simple} report the result for \texttt{in-2004} ($\mathcal{Q}= \numprint{0.980}$) which is comparable to the result that we report here.
D{\v{z}}ami{\'{c}}~\cite{Dzamic2017} build upon VNS proposing an ascent-decent VNS. 
Results are reported for \texttt{celegans\_metabolic} ($\mathcal{Q} = 0.453248$), \texttt{email} ($\mathcal{Q}=\numprint{0,582829}$), \texttt{polblogs} ($\mathcal{Q} = \numprint{0.427105}$) for which their algorithm computes the same results as all other tools reported in Table~\ref{fig:modularityscores}.
Moreover, the following best results are reported \texttt{as-22july06} ($\mathcal{Q}=\numprint{0.678381}$), \texttt{astro-ph} ($\mathcal{Q} = \numprint{0.745246}$), \texttt{cond-mat-2005} ($\mathcal{Q}=\numprint{0.747181}$), \texttt{PGPgiantcompo} ($\mathcal{Q}=0.886647$), as well as \texttt{power} ($\mathcal{Q}=0.940974$) which are indeed better than the previous best result obtained during the 10th DIMACS challenge, but still worse than average result of our algorithm in every case.
Hence, overall we consider our algorithm as a new state-of-the-art heuristic for solving the modularity clustering problem.

\subparagraph*{Convergence Plots.}
We are now interested in how solution quality evolves over time.
We start by explaining how to compute the data for a single instance $I$, i.e., a clustering of a graph $G$.
Whenever an algorithm computes a clustering that \emph{improves} the modularity score, it
reports a pair ($t$, $\mathcal{Q}$), where the timestamp $t$ is the currently elapsed time and $\mathcal{Q}$ is the modularity score of the currently reported solution.
For $r$ repetitions with different seeds $s$, these $r$ sequences $T_s^I$ of
pairs are merged into one sequence  $T^I$ of triples $(t, s, \mathcal{Q})$, which is sorted by the timestamp $t$.
Since we are interested in the \emph{evolution} of the solution quality, we compute the sequence $T_{\text{max}}^I$ representing \emph{event-based} average values.
We start by computing the average objective value $\overline{\mathcal{Q}}$ and the average time $\overline{t}$ using the first pair ($t$, $\mathcal{Q}$) of all $r$ sequences $T_s^I$ and insert
$(\overline{t}, \overline{\mathcal{Q}})$ into $T_{\text{max}}^I$. We then sweep through the remaining entries $(t, s, \mathcal{Q})$ of $T^I$. Each entry corresponds
to a clustering computed at timestamp $t$ using seed $s$ that improved the solution quality to $\mathcal{Q}$. For each entry we replace the old objective value of seed $s$
that took part in the computation of $\overline{\mathcal{Q}}$ with the new value $\mathcal{Q}$, recompute $\overline{\mathcal{Q}}$ and insert a new pair $(t, \overline{\mathcal{Q}})$ into $T_{\text{max}}^I$.
$T_{\text{max}}^I$ therefore represents the evolution of the average solution quality $\overline{\mathcal{Q}}$ for instance $I$ over time.

We present those plots for a selected set of instances in Figure~\ref{fig:convergenceplots}. In the plots one can see that solution quality levels off in the beginning. At this point, the algorithm is still in the initialization phase, where the population is built using repeated runs of the Louvain method with different coarsening strategies. However, this is not enough to find significantly improved solutions. Afterwards, solution quality improves rapidly as the actual evolutionary operations run.
\renewcommand{\arraystretch}{0.6}
\section{Conclusion}
\label{conclusion}
We presented a parallel memetic algorithm, VieClus, that tackles the graph clustering problem. 
A key component of our contribution are natural recombine operators that employ ensemble clusterings as well as multi-level techniques. We combine these techniques with a scalable communication protocol, producing a system that is able to reproduce or improve previous all entries of the 10th DIMACS implementation challenge under consideration as well as results recently reported in the literature in a short amount of time.
Moreover, while the previous best result for different instances has been computed by a variety of solvers, our algorithm can now be used as a single tool to compute the result.
Hence, overall we consider our algorithm as a new state-of-the-art heuristic for solving the modularity clustering problem.
Considering the good results of our algorithm, we want to further improve and release it.
In the future, it may be interesting to instantiate our scheme for different objective functions depending on the application domain or to use a more diverse set of initial solvers to create the population. We also want to look at distributed memory parallel multi-level algorithms for the problem that can use the depicted algorithm as an initial clustering scheme on the~coarsest~level~of~the~hierarchy.

\section*{Acknowledgements}
The research leading to these results has received funding from the European Research Council under the European Community's Seventh Framework Programme (FP7/2007-2013) /ERC grant agreement No. 340506.

\vfill
\pagebreak
\bibliography{phdthesiscs}
\vfill
\pagebreak
\begin{appendix}
\section{Approximate Bound for Modularity}
The modularity clustering problems seeks to find a clusterings such that cov($\mathcal{C}$) - $\mathbb{E}[$cov($\mathcal{C}$)] is maximum.
However, the problem is NP-complete and our algorithm heuristically finds solutions. 
Moreover, modularity is trivially bounded by 1. The true optimum clustering, however, has typically a value below that. Thus we try to find a bound or a value to normalize the score such that one gets a clearer picture on how far away from the optimum score the achieved modularity value is.

To do so, let us define $ \sum_{v \in V_i} d(v)$ to be the volume vol($V_i$) of the cluster $V_i$.
Note that the formula 
$\frac{1}{4m^2} \sum_{V_i \in \mathcal{C}} \text{vol}(V_i)^2$ is minimized if all clusters have the same volume. 
But again, finding the optimum value of this equation is hard.
As $\sum_{V_i \in \mathcal{C}} \text{vol}(V_i) = 2m$, the equation is minimized when vol($V_i$)$=2m/k$ $\forall i = 1 \ldots k$, where $k$ is the number of clusters. Thus it follows that $\frac{1}{4m^2} \sum_{V_i \in \mathcal{C}} \text{vol}(V_i)^2 \geq \frac{1}{4m^2} \cdot k \cdot (\frac{2m}{k})^2 = \frac{1}{k}$. This gives an upper of $1-\frac{1}{k}$ for $\mathcal{Q}(\mathcal{C})$ if the number of clusters is an input to the clustering problem. This upper bound is, however, not very far from the trivial upper bound~of~1~for~$k > 10$. 

Thus for each graph we also clustered the nodes into cluster $V_i$ using the following heuristic to try to balance the volumes of the resulting clusters as much as possible: We sort the nodes in decreasing order of their degree and then assign them step by step to the cluster having the smallest volume. We break ties by using the cluster with the smallest ID among those having the smallest volume. For this heuristic, we use the number of clusters that our algorithm has computed as value for $k$. For the resulting clustering $\mathcal{C}^*$ we compute  the value $\mathbb{E}[\text{cov}(\mathcal{C}^*)]$ and use $B := 1-\mathbb{E}[\text{cov}(\mathcal{C}^*)]$ as a value to normalize $\mathcal{Q}(\mathcal{C})$. Table~\ref{tab:bound} shows the resulting bound and compares them against our results.
\begin{table}[h!]   
\begin{center}
\begin{footnotesize}
\caption{Columns from left to right: the best score computed of our evolutionary algorithm, the approximate bound computed with the heuristic algorithm to balance cluster volumes and the ratio of the two values.}
\label{tab:bound}
\begin{tabular}{|l||l|l||l|}
\hline
Graph & Max. $\mathcal{Q}$&  $B$ & $\mathcal{Q}/B$ \\
\hline
\hline
\multicolumn{4}{|c|}{Small Instances} \\
\hline
\hline
\texttt{as-22july06}                                                 & \numprint{0.679396} &\numprint{0.973684}&\numprint{0.697758}\\
\texttt{astro-ph}                                                    & \numprint{0.746292} &\numprint{0.999070}&\numprint{0.746987}\\
\texttt{celegans\_metabol}                                           & \numprint{0.453248} &\numprint{0.888889}&\numprint{0.509904}\\
\texttt{cond-mat-2005}                                               & \numprint{0.750171} &\numprint{0.999468}&\numprint{0.750570}\\
\texttt{email}                                                       & \numprint{0.582829} &\numprint{0.900000}&\numprint{0.647588}\\
\texttt{PGPgiantcompo}                                               & \numprint{0.886853} &\numprint{0.989796}&\numprint{0.895996}\\
\texttt{polblogs}                                                    & \numprint{0.427105} &\numprint{0.996168}&\numprint{0.428748}\\
\texttt{power}                                                       & \numprint{0.940977} &\numprint{0.975610}&\numprint{0.964501}\\
\texttt{smallworld}                                                  & \numprint{0.793187} &\numprint{0.995652}&\numprint{0.796651}\\
\texttt{memplus}                                                     & \numprint{0.701275} &\numprint{0.984127}&\numprint{0.712586}\\
\texttt{G\_n\_pin\_pout}                                             & \numprint{0.500466} &\numprint{0.993865}&\numprint{0.503555}\\
\texttt{caidaRouterLevel}                                            & \numprint{0.872828} &\numprint{0.997758}&\numprint{0.874789}\\
\texttt{rgg\_n17}                                                    & \numprint{0.978454} &\numprint{0.992537}&\numprint{0.985811}\\
\texttt{luxembourg.osm}                                              & \numprint{0.989672} &\numprint{0.996283}&\numprint{0.993364}\\
\hline                                                                                                         
\hline                                                                                      
\multicolumn{4}{|c|}{Large Instances}\\
\hline
\hline
\texttt{coAuthorsCiteseer}           &   \numprint{0.906830} & \numprint{0.995575}&\numprint{0.910861}\\
\texttt{citationCiteseer}            &   \numprint{0.825545} & \numprint{0.993289}&\numprint{0.831123}\\
\texttt{coPapersDBLP}                &   \numprint{0.868058} & \numprint{0.995283}&\numprint{0.872172}\\
\texttt{belgium.osm}                 &   \numprint{0.995064} & \numprint{0.998358}&\numprint{0.996701}\\
\texttt{ldoor}                       &   \numprint{0.970555} & \numprint{0.989796}&\numprint{0.980561}\\
\texttt{eu-2005}                     &   \numprint{0.941575} & \numprint{0.996564}&\numprint{0.944821}\\
\texttt{in-2004}                     &   \numprint{0.980690} & \numprint{0.999136}&\numprint{0.981538}\\
\texttt{333SP}                       &   \numprint{0.989356} & \numprint{0.995726}&\numprint{0.993603}\\
\texttt{prefAttachment}              &   \numprint{0.316089} & \numprint{0.875000}&\numprint{0.361245}\\
\hline                                                                            
\end{tabular}
\end{footnotesize}
\end{center}
\vspace*{-.5cm}
\end{table}
\vfill
\pagebreak

\end{appendix}
\end{document}